# Design and Performance Evaluation of an Elbow-Based Biomechanical Energy Harvester

**Hubert Huang[1], Jeffrey Huang[2]**
[1]Department of Computer Science, Pacific American School, Zhubei City, Hsinchu 30272 Taiwan
[2]Department of Computer Science, Columbia University, New York, NY 10027 USA

Corresponding author: Hubert Huang (e-mail: huberthuang0930@gmail.com).

**ABSTRACT** Carbon emissions have long been attributed to the increase in climate change. With the effects of climate change escalating in the past few years, there has been an increased effort to find green alternatives to power generation, which has been a major contributor to carbon emissions. One prominent way that has arisen is biomechanical energy, or harvesting energy based on natural human movement. This study will evaluate the feasibility of electric generation using a gear and generator-based biomechanical energy harvester in the elbow joint. The joint was chosen using kinetic arm analysis through MediaPipe, in which the elbow joint showed much higher angular velocity during walking, thus showing more potential as a place to construct the harvester. Leg joints were excluded to not obstruct daily movement. The gear and generator type was decided to maximize energy production in the elbow joint. The device was constructed using a gearbox and a generator. The results show that it generated as much as 0.16 watts using the optimal resistance. This demonstrates the feasibility of electric generation with an elbow joint gear and generator-type biomechanical energy harvester.

**INDEX TERMS**    Biomechanical Energy, Machine Learning Kinetic Analysis

## I. INTRODUCTION

As global warming intensifies, a pressing issue during the present G20 summit was the goal to reduce carbon dioxide emissions, a discussion prompted by record-breaking temperatures and climate catastrophes [1]. One of the well-documented reasons behind the rapid climate change is carbon dioxide emissions. A study by Solomon et al.[2] as early as 2009 has emphasized the irreversible change that our heightened carbon emissions have on climate change. As mentioned by the US Environmental Protection Agency, around 25 percent of current carbon emissions are due to energy consumption, usually through fossil fuels [3]. A study by Zhang et al.[4] echoes this sentiment through a case study of Chinese energy consumption and carbon emissions, which finds a positive correlation between increased use of nonrenewable fuels and carbon emissions. In response to this increasing crisis, most countries signed the Paris Agreement that strived to limit carbon emissions through ways such as transitioning to renewable energies [5]. Thus, the world has been headed in a direction that favors more green sustainable energy that leads to fewer negative effects on the environment [6].

One rising form of sustainable energy production that has gained traction is biomechanical energy [7]. According to Choi et al.[8] in a study published by the Journal of Energies, biomechanical energy, or the energy harvested from natural human movements, has risen in importance as an effective kinetic energy harvesting method. The study listed several prominent biomechanical energy harvesting methods like the Piezoelectric Energy harvester using small vibrations, the Triboelectric generator using nano electric potential differences, and the Electromagnetic energy harvester using a typical generator all as viable ways to generate energy based on natural human movement.

Similar to the wide array of different ways to harvest biomechanical energy, there has also been extensive literature documenting the possible areas in which biomechanical energy can be harvested. Niu et al. [9]document the possible energy that can be produced from each joint, with arm, elbow, and leg joints producing sizable amounts of power.

While there are abundant sources that analyze biomechanical energy for piezoelectric, triboelectric energy harvesters, or electromagnetic energy harvesters on leg joints, there still remains to be few literature discussing energy generated in the arm joints, especially the elbow and shoulder joints. Thus, this study will evaluate the feasibility of a gear and generator energy harvester by harvesting energy from the elbow joint.

## II. Literature Review

### A. *Potential biomechanical energy Harvesting*

In the human body, several significant areas produce sizable power that can be harvested. The aforementioned study by Niu et al. [9] points out that heel strike, ankle, knee, hip, elbow, and shoulder joints are all potential areas where biomechanical energy harvesting can be conducted. The study points out that the hip, ankle, and knee joints all have significantly higher power (over 30 watts). Another study by Li et al. reaffirms the applicability of using knee joints to harvest biomechanical energy [10]. The study states that their biomechanical harvester can generate up to as much as $4.8 \pm 0.8$ W without affecting metabolism and $5.6 \pm 21$ W if the person's movement is affected. Using the knee joint remains to be one of the most common and efficient biomechanical harvesters.

### B. *Current Biomechanical Energy Technologies*

Just as Choi et al. mention, there are generally four distinct types of biomechanical generator methods [8]. These are the Piezoelectric energy harvester, Triboelectric energy harvester, Inertial Inductor type of Electromagnetic energy harvester, and Gear and Generator Type of Electromagnetic energy harvester. Each one has its own pros and cons as well as different applications that are suitable for it.

The first distinct energy harvesting method is the Piezoelectric energy harvester. The harvester works with the piezoelectric effect, which is the ability of materials to generate electric charges during vibrations. According to the aforementioned Choi et al. [8], the average piezoelectric energy harvester is used on the knee or foot strike, which allows for vibrations, either from the movement of the knee joint or the force of a foot stepping, that thus allows for the piezoelectric effect generating energy in the mW range. A significant benefit of the piezoelectric effect energy generator is that it is small and can be conveniently placed in multiple places, but can also be fragile. [11]

The second distinct method is the triboelectric energy harvester. The harvester uses the triboelectric effect, which is an electric transfer effect that generates electric charge through friction [12]. This is most commonly used in footstrike scenarios, where there is a high amount of friction and vibrations that are beneficial for generating electric current. However, according to Yang et al [13], since the frequency of walking remains relatively low, the power generated by triboelectric energy harvesters is a limiting factor. Moreover, the study mentions that triboelectric generators are still inefficient, as most of the energy from these footstrikes dissipates as heat.

The third mentioned distinct method is the inertial induction harvester of electromagnetic energy. As mentioned by Choi et al.[8], this method includes rotating a permanent magnet or coil with a rotating coil around. There needs to be maximized linear velocity and oftentimes a coil or spring. The pros of the inertial induction harvester are that it minimally interferes with human movement, but the energy generated is still in the mW range, which is on the lower end.

The final method is the gear and generator way of harvesting electromagnetic energy. This is the most straightforward method of biomechanical energy harvesting, using a conventional generator. According to Choi et al. [8], this type of generator can generate up to as much as the W level, which is the highest out of the other biomechanical energy harvesters.

Gurusamy et al.[14] show the potential of the electromagnetic generator harvesting method, as they mention how the electromagnetic harvester could generate up to 5.09 W and 14.95 W. This once again shows the inherent power advantage that an electromagnetic generator has. Along with this heightened energy production, there are concerns that it affects human movement more than the other generators that generate less electricity. Wu et al. [15] mention the inevitable metabolism cost of these devices on the knee, which accounts for 2.3 percent of total metabolism. However, the study also mentions that these costs are usually negligible. This shows how the metabolism cost is something to consider for gear and generator models in joints where movement is important.

## III. Methodology

### A. *Deciding the area of development*

Before deciding on the specific designs of the mechanical arm, the specific area of the body from which energy is harvested has to be decided first. From the previously mentioned literature, Niu et al. [9] suggest that areas such as shoulder, elbow, and knee joints all generate significant amounts of biomechanical energy. In this study, one specific area would be chosen to generate biomechanical energy. The two main areas of consideration are the knee and the arm, which includes both the shoulder and the elbow. As mentioned previously by Li et al. [10], when a knee biomechanical energy harvester has little metabolism cost, or the amount of extra energy by the human body to power the harvester, it generates significantly lower wattage. While multiple studies have attempted to decrease the metabolism cost of knee harvesters to decrease its effect on walking, there will always be a metabolism cost because the knee is a fundamental part of walking and human movement [16]. Thus due to the potential issues that generating power with the knee causes, this study will focus on the arm area.

Thus, this study will now decide between using the elbow and the arm joint. In order to analyze which area has more potential to be harvested, this study will both conduct its own experiments as well as analyze previous literature.

As previously mentioned by Niu et al. [9], the elbow and shoulder joint have approximately the same wattage and power of generation, with the elbow having 2.2 watts and the shoulder having 2.1 watts. Thus, this study will decide between these 2 areas to develop the biomechanical arm using motion tracking, figuring out which area has more potential to generate power.

In order to analyze gait and movement dynamics, this study will track human poses. This study will use the method outlined in Gupta et al. [16]. While the study is mainly focused on tracking knee flexion and gait, this study will use the same methods employed but on arm and shoulder dynamics. Gupta et al. [16] mainly outline 2 types of motion tracking: Kinovea, a marker-based motion tracking software, and MediaPipe, a computer-vision-based machine learning tracking software. This study affirms the practicality of using MediaPipe for pose analysis, having as high as 0.941 correlation with Kinovea. While Kinovea needed markers, Gupta et al.[16] confirm the use of MediaPipe Pose analysis, which is markerless and more flexible in this situation. Singh et al. [17] further emphasize the practicality of using MediaPipe for pose analysis, noting that it can provide up to 501 landmarks in its current model. Thus, due to the flexibility that MediaPipe offers with its markerless tracking, this study will employ it for motion analysis.

The main objective of this section is to analyze the angular speed and movement of each of the joints in walking and running in order to find the most effective area to create the biomechanical device. For motion tracking, this study will base the sample on a typical walking movement by a healthy teenager. The motion tracking will be conducted using a machine learning-based MediaPipe Body landmark system.

Firstly, the study recorded a video of the person walking in 3 separate speeds: 4km/hr, 6km/hr, and 8km/hr. Then, the study will record the same video at these 3 speeds but the person is running instead of walking.

After recording the videos, this study implemented 2 main types of landmark tracking at every speed and video. First, the elbow and wrist landmarks are tracked to determine the elbow angle cycle. Treating the wrist landmark as a "fixed point," the code tracks the change in the angle of the elbow joint. The same is done with the shoulder joint. Using the trunk or shoulder joint and elbow joint landmark, the shoulder joint angle can also be tracked.

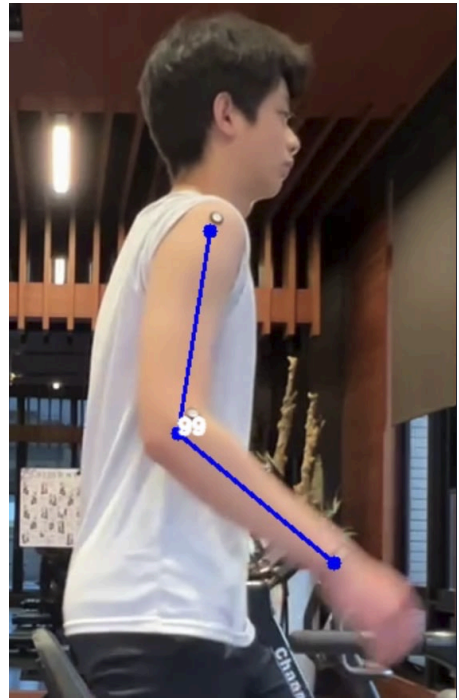

FIGURE 1. Sample of MediaPipe instantaneous joint tracking and angle calculation

After calculating the instantaneous angular velocity for the joints, a plot of the angular velocity was plotted out, as well as the calculation for the root mean square of the angular velocity on each joint. The same process was conducted for 2 separate case studies to consolidate the findings.

These are the results of the first participant, with each of the numbers in the unit of radians per second. To better illustrate these findings, a graph of the respective angular velocities was given to show the general trend.

TABLE I
FIRST PARTICIPANT JOINT ANALYSIS

| Categories | Walking Elbow | Walking Trunk | Running Elbow | Running Trunk |
|---|---|---|---|---|
| Speed | | | | |
| 4km/hr | 1.5 | 1.1 | 1.9 | 3.1 |
| 5km/hr | 3.8 | 0.6 | 1 | 2 |
| 6km/hr | 7.1 | 0.6 | 0.9 | 2.3 |

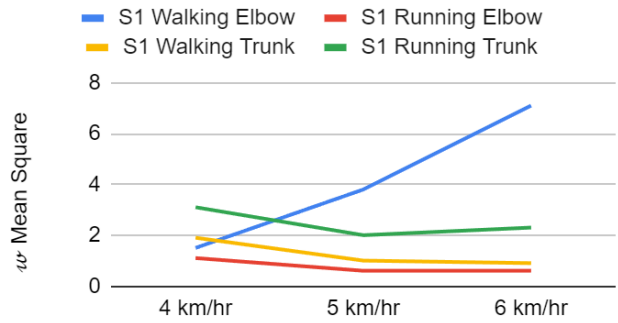

FIGURE 2.  **Summarized results of the kinetic analysis on the first case study.**

TABLE II
SECOND PARTICIPANT JOINT ANALYSIS

| Categories | Walking Elbow | Walking Trunk | Running Elbow | Running Trunk |
|---|---|---|---|---|
| Speed | | | | |
| 4km/hr | 8.9 | 3.6 | 1.9 | 5.2 |
| 5km/hr | 11 | 5.6 | 1 | 9.6 |
| 6km/hr | 9.4 | 7.1 | 0.9 | 10.1 |

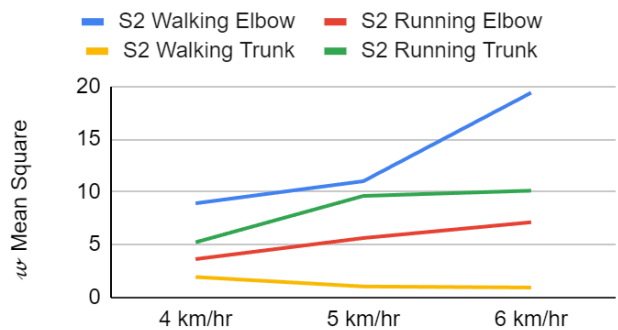

FIGURE 3.  **Summarized results of the kinetic analysis on the first case study.**

As it is apparent on both of the charts, the blue line, which represents the angular speed of the elbow joint during walking, is highest by a margin, especially as the walking speed is increased. Thus, this study will model the biomechanical device to maximize the energy gathered by the elbow joint during walking.

### B. *Method of Electric Generation*

As previously stated in the literature review, there are four main types of biomechanical harvesting methods, The electromagnetic energy harvester with gear and generator was selected due to the study also choosing the elbow joint. To maximize energy harvested, the gear and generator would be the most suitable, as an arm would be able to power the generator amplified by the gears, thus producing the most energy. While it is possible to use other methods like the induction type of electromagnetic or triboelectric harvester, the generating power from a gear and generator type is much higher, thus making it a lot more feasible.

### C. *Designing the Biomechanical Arm*

After deciding that this study will use the gear and generator harvester to harvest energy from the ankle, this study will thus start designing the structure of the biomechanical arm. The arm is largely split into 3 main components: the "arm", the gearbox, and the generator.

1) DESIGNING THE GEARBOX

A staple of a biomechanical energy generator is the gearbox. The gearbox is meant to enhance the power generated by the motor by amplifying the gear ratio.

The first and most important aspect of the biomechanical arm is the gearbox. For designing the gearbox, it is important to strike a balance between maximizing the overall gear ratio to increase electric power generation and lowering the gear ratio enough so that a human arm in normal walking movement can easily turn the generator.

Before deciding the specific gear ratios of the gearbox, it is important to decide the specific gear used. For this study, in order to minimize the weight of the gearbox, we will be using double-sided spur gears for the main gearbox. A double-sided gear has a smaller gear on one side and a bigger gear on another side. The benefit of double-sided spur gear is that it would be more cohesive and less complicated to create a gear train.

After choosing the type of gear, the study will thus design the specific gear ratios. All the gears have to be the same module M to be used effectively in a gear train. Balancing between higher gear ratios that have higher resistive torque and lower gear ratios that have less electric generation, this study decided to use the following gear train to achieve a gear ratio of 27.2.

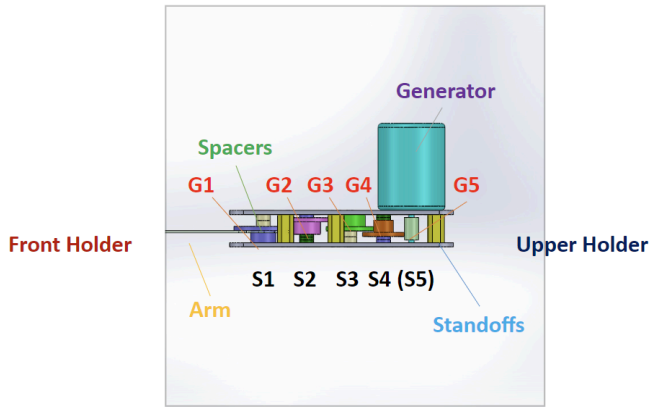

FIGURE 4.  Design graph or picture of the gearbox on the Biomechanical arm

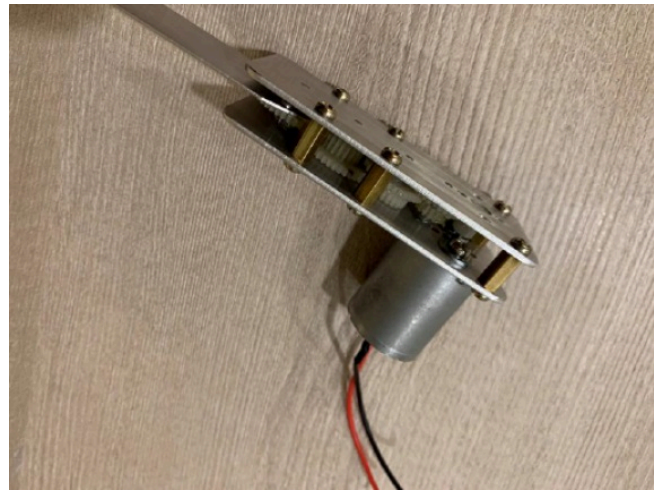

FIGURE 5.  Picture of the final completed gearbox for the biomechanical arm

TABLE III
GEARBOX GEAR RATIOS AND DETAILS

| Gears | G1 | G2 | G3 | G4 | G5 |
| --- | --- | --- | --- | --- | --- |
| Gear Module | 0.5 | 0.5 | 0.5 | 0.5 | 0.5 |
| Number of Gear Teeth (Large Gear) | 42 | 40 | 32 | 30 | 10 |
| Number of Gear Teeth (Small Gear) | 18 | 20 | 22 | 14 | 10 |
| Pitch Diameter (Big gear) | 21 | 20 | 17 | 15 | 5 |
| Pitch Diameter (Small Gear) | 9 | 10 | 11 | 7 | 5 |
| Center Distance(mm) | N/A | 15.5 | 15.5 | 12 | 10 |

After creating the gearbox, a front holder was connected to the generator, and the upper arm and a shaft with a back holder were connected to the other end of the gear train and the lower arm, which completed the entire design.

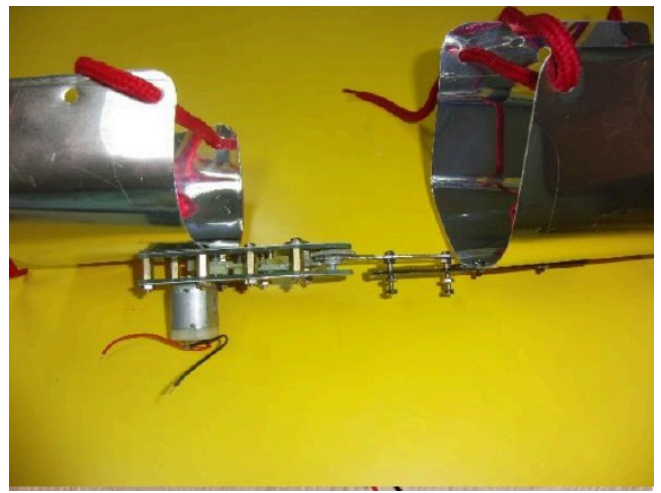

FIGURE 6.  Picture of the final completed prototype of the biomechanical arm

2) DERIVING THE EQUATIONS

After designing the basics of the mechanical arm, it is also important to the study to analyze and derive the equations that govern the physics of the biomechanical arm.
From the basics of physics, the study can first derive the basic formula for power.
The following derivations are all based on derivations made by Li et al.[10].

$$P_a = M_a \times \omega_a$$

The $P_a$ here represents the power of the arm while Ma is the mass of the arm and Wa is the rotational speed of the arm.

Moreover, in constructing the mechanical arm, a gear ratio $r_t$ will be applied. Thus the final $\omega_g$ of the arm encompassing the gear ratios will be:

$$\omega_g = \omega_a \cdot r_t$$

After getting the baes angular speed that the motor would run on, it is then thus possible to get the voltage created by the generator. The voltage by a generator is directly proportional to the angular speed after the gear ratio as well as the K constant of the motor. Thus the equation for the voltage is represented below:

$$V = K_g \omega_g$$

The K constant correlates with both 3 main factors inside a generator.
1. The number of coils in a generator
2. The magnetic field of the permanent magnet inside
3. The area of the coils

For the voltage itself, there is also another form using the currents that will be inside the arm.

$$V = I \cdot R_g + I \cdot R_l$$

$R_g$ represents the internal resistance of the generator, while $R_l$ represents the resistance of the extra load that is added to the generator. Thus, combining the previous 2 equations for the voltage, the following equation can be derived for the current of the mechanical arm

$$V = K_g \omega_g = I \cdot R_g + I \cdot R_l$$

$$I = \frac{K_g \omega_g}{R_g + R_l} = \frac{K_g \cdot r_t}{R_g + R_l} \cdot \omega_a$$

After getting the current, the equation for power can be thus derived. The base equation for power can be shown below.

$$P_o = V_l^2 \div R_l = I^2 \cdot R_l$$

Using the previously current formula, we can get the power dissipated formula of

$$P_o = \left(\frac{K_g \cdot r_t}{R_g + R_l} \cdot \omega_a\right)^2 \cdot R_l$$

However, the study can't simply study the power from the resistance of the load. The total dissipated power from the electric generation should also account for the power dissipated from the internal resistance. Accounting for the internal resistance, the total power is represented by the following equation.

$$P_g = I^2 \cdot R_g$$
$$P_a = P_o + P_g$$

Thus, if this study wants to calculate the efficiency of the power generation of the arm. All we need to do is divide the power dissipated by the load by the total power, which is the power dissipated by the load and the internal resistance. The equation is shown below

$$Efficiency = P_o \div P_a$$

## IV. Results and Discussion

After the construction of the device, this study therefore put the device to the test. However, in accordance with the function previously, it is clear that the resistance of the load is proportional to the total resistance, but it is inversely proportional to the current. Thus, it is important to find a good resistance load to maximize the power generated. Therefore there was a wide array of external resistance loads that was tested and the power generated for each one was calculated.

| Resister (ohm) | RMS (v) | Power (w) |
|---|---|---|
| 5 | 0.72 | 0.10 |
| 6.8 | 0.85 | 0.11 |
| 8.2 | 1.01 | 0.12 |
| **9.1** | **1.2** | **0.16** |
| 10 | 1.2 | 0.14 |
| 14.3 | 1.35 | 0.13 |
| 20 | 1.6 | 0.13 |

FIGURE 7. Table of the generated power versus the placed internal resistance

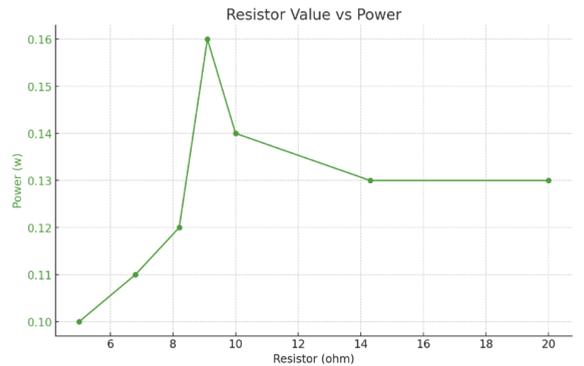

FIGURE 8. Graph of the generated power versus the placed internal resistance

After testing varying resistances, the study settled that the resistance of 9.1 ohms would lead to the maximum amount of power generated, as any resistances tested above and below yielded less power than it.

| Arm Swing degree | Time (min) | Capacitor | Measured Voltage |
|---|---|---|---|
| 45~65 | 15 | 25F/5.4V | 0.642V |
| 65~90 | 15 | 25F/5.4V | 1.117V |

FIGURE 9. Table of measured voltage after time span

Once again using the final configuration, the final product delivered an average voltage of 0.6425 volts if the elbow swing was around 45 to 65 degrees, and a much higher 1.117 volts when the elbow swing was from 65 to 90 degrees.

**CONCLUSION**

This study demonstrates the feasibility of constructing a biomechanical arm using a gear and generator harvester type to generate energy from the elbow joint. The decision comes after an analysis using a machine-learning-based MediaPipe algorithm decided that the elbow joint, compared to the shoulder joint, offered a higher amount of potential, especially during walking, as the angular speed was much higher than the shoulder joint. After deciding on the specific joint, the study also decided that the gear and generator type of harvester would be the most suitable because it has by far the highest power-generating capacity. After deciding on these, this study designed the biomechanical arm with a generator and a gearbox, amplifying the gear ratio to increase the electricity generated. After the design, the study also tested the electricity-generating capacities of the arm, concluding that the external resistance of 9.1 ohms was ideal for maximizing the power generated. The device generated approximately 0.642 to 1.117 volts during normal usage. This study once again demonstrates the feasibility of using a biomechanical energy harvester. Future directions could be made to increase the gear ratio or use another harvester method to try to produce energy in the same elbow joint.